\documentclass[runningheads]{llncs}

\usepackage[mobile]{eccv}

\usepackage{eccvabbrv}

\usepackage{graphicx}
\usepackage{booktabs}
\usepackage{caption}     
\usepackage{makecell}
\usepackage{wrapfig}
\usepackage[accsupp]{axessibility}  

\usepackage{booktabs}   
\usepackage{multirow}   
\usepackage{amsmath}
\usepackage{makecell} 
\usepackage{graphicx}   
\usepackage{balance}
\usepackage[table]{xcolor} 
\usepackage{tabularx} 
\usepackage[bottom]{footmisc}
\usepackage[ruled,vlined,linesnumbered]{algorithm2e}
\usepackage{algpseudocode}
\usepackage{authblk}
\input{insbox}
\definecolor{bestcolor}{HTML}{C8E6C9}     
\definecolor{secondbestcolor}{HTML}{E8F5E9} 
\definecolor{thirdbestcolor}{HTML}{F1F8E9}  

\usepackage{hyperref}

\usepackage{orcidlink}

\begin{document}

\title{RoboStereo: Dual-Tower 4D Embodied World Models for Unified Policy Optimization} 

\titlerunning{RoboStereo}


\author{%
  Ruicheng Zhang\inst{1}\thanks{These authors contributed equally to this work.} \and
  Guangyu Chen\inst{1}$^*$ \and
  Zunnan Xu\inst{1,2}\thanks{Project Lead.} \and
  Zihao Liu\inst{1} \and
  Zhizhou Zhong\inst{3} \and
  Mingyang Zhang\inst{1} \and 
  Jun Zhou\inst{1}$^\ddagger$ \and 
  Xiu Li\inst{1}\thanks{Corresponding author.}
}

\authorrunning{R. Zhang et al.}

\institute{%
  $^1$Tsinghua University, \quad
  $^2$X Square Robot, \quad 
  $^3$HKUST
}

\maketitle

\begin{abstract}
Scalable Embodied AI faces fundamental constraints due to prohibitive costs and safety risks of real-world interaction. While Embodied World Models (EWMs) offer promise through imagined rollouts, existing approaches suffer from geometric hallucinations and lack unified optimization frameworks for practical policy improvement. We introduce RoboStereo, a symmetric dual-tower 4D world model that employs bidirectional cross-modal enhancement to ensure spatiotemporal geometric consistency and alleviate physics hallucinations. Building upon this high-fidelity 4D simulator, we present the first unified framework for world-model-based policy optimization: (1) Test-Time Policy Augmentation (TTPA) for pre-execution verification, (2) Imitative-Evolutionary Policy Learning (IEPL) leveraging visual perceptual rewards to learn from expert demonstrations, and (3) Open-Exploration Policy Learning (OEPL) enabling autonomous skill discovery and self-correction. Comprehensive experiments demonstrate RoboStereo achieves state-of-the-art generation quality, with our unified framework delivering >97\% average relative improvement on fine-grained manipulation tasks.
\end{abstract}

\section{Introduction}
\label{sec:intro}
Vision-Language-Action (VLA)~\cite{vlarft,robridge,pi0,zhai2025igniting,a0} models represent a promising par-adigm for general-purpose robotic manipulation, allowing robots to interpret open-ended instructions and execute diverse tasks. Unlike conventional language models that process static text, robotic policies must operate in continuous, dynamic environments with physically realistic feedback. However, scaling such models in the physical world encounters fundamental barriers that persist across both training and inference. Real-world interaction remains prohibitively expensive, slow, and inherently unsafe, severely constraining the collection of large-scale, diverse training data and impeding safe action verification and refinement at deployment time~\cite{WorldArena}.

\begin{figure*}[t]
  \centering
  \setlength{\abovecaptionskip}{-0.02em}   
   \includegraphics[width=1.0\linewidth]{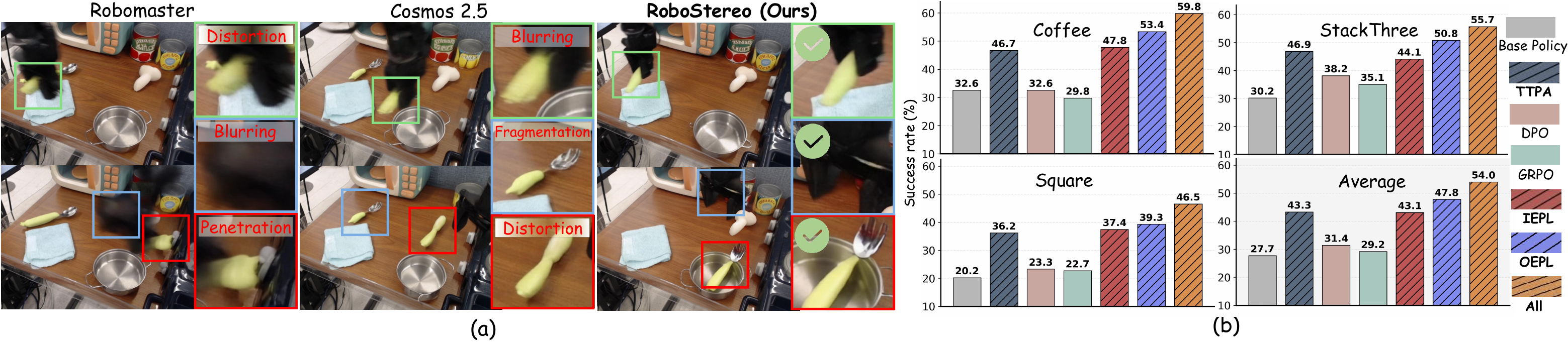}

  \caption{
(a) Qualitative comparison of RoboStereo against SOTA EWMs. (b) Quantitative comparison of unified policy optimization framework against traditional paradigms.
} 
\label{fig:vis_con}
   \vspace{-2em}
\end{figure*}

To overcome the physical-world constraints, Embodied World Models (EWMs) ~\cite{gigaworld,genie,mind,robomaster,vidarc,cosmos,IRASim,wow} have emerged as a promising alternative. By learning to predict future observations conditioned on robot actions, EWMs act as differentiable digital twins, enabling agents to plan, verify, and optimize policies entirely within safe, imagined rollouts. Despite this potential, existing EWMs face two fundamental limitations that hinder robust policy optimization in practice. First, current EWMs struggle to maintain 3D geometric consistency in complex interaction modeling~\cite{one4d}. Despite producing visually appealing frames, they suffer from physics hallucinations (e.g., object teleportation, scale drift, surface penetration, Fig.~\ref{fig:vis_con}(a)) due to the absence of explicit geometric grounding. Second, although isolated prior works~\cite{WMPO,vlarft,TGRPO2025} have explored world-model-based policy improvement, no unified framework exists capable of leveraging high-fidelity EWMs for the full spectrum of policy optimization from inference-time imagined verification to training-time evolutionary refinement.

In this work, we introduce RoboStereo, a dual-tower 4D world model designed for unified policy optimization across training and inference. To enforce geometric awareness and 3D consistency, RoboStereo employs a symmetric twin backbone Diffusion Transformer (DiT)~\cite{DiT} architecture that processes RGB videos and 3D pointmaps as mutually reinforcing modalities. Through bidirectional cross-attention across the two towers, the video branch incorporates geometric constraints to maintain structural rigidity, while the pointmap branch leverages semantic context to refine object geometry. This symmetric fusion grounds every generated pixel in consistent 3D space, yielding temporally coherent 4D point-cloud trajectories. A dedicated 4D Gaussian Splatting head~\cite{3Dgaussians} then renders these trajectories into photorealistic observations from flexible viewpoints, supporting multi-view supervision during policy learning.

Building on RoboStereo as a high-fidelity 4D mental simulator, we introduce the first comprehensive framework for world-model-based policy optimization. It comprises three complementary paradigms that span the full spectrum from inference-time improvement to training-time refinement: 
(1) Test-Time Policy Augmentation (TTPA), a zero-shot pre-execution mechanism that validates and refines candidate action sequences through imagined rollouts; 
(2) Imitative-Evolutionary Policy Learning (IEPL), which leverages dense visual-imitation rewards from flexible-view imagined trajectories to efficiently align policies with expert behavior; 
(3) Open-Exploration Policy Learning (OEPL), a self-supervised approach that enables autonomous skill discovery and self-correction in the absence of expert demonstrations.
This framework is the first to treat an EWM as the foundational backbone for scalable VLA policy improvement across both deployment and learning phases, substantially decoupling optimization from expensive and risky physical interaction.

Our main contributions are as follows:

\begin{itemize}
\vspace{-1em}
    \item We propose RoboStereo, a symmetric dual-tower Diffusion Transformer (DiT) that jointly processes RGB video and 3D pointmap sequences as complementary modalities. Bidirectional cross-attention enforces strict spatiotemporal consistency, yielding high-fidelity 4D generation grounded in explicit geometric constraints.
    
    \item We introduce the first unified framework for world-model-based policy optimization in VLA settings, encompassing three paradigms: TTPA for pre-execution validation and refinement, IEPL for viewpoint-flexible visual imitation and evolutionary refinement, and OEPL for demonstration-free skill discovery and self-correction.
    
    \item Extensive experiments show that RoboStereo achieves state-of-the-art performance in physics adherence and 3D geometric accuracy. Moreover, the unified optimization framework delivers substantial policy gains (over 97\% mean relative improvement over the baseline) on fine-grained manipulation tasks, demonstrating the effectiveness of world model-based policy improvement.
\end{itemize}

\begin{figure*}[t]
  \centering
  \setlength{\abovecaptionskip}{-0.02em}   
   \includegraphics[width=1.0\linewidth]{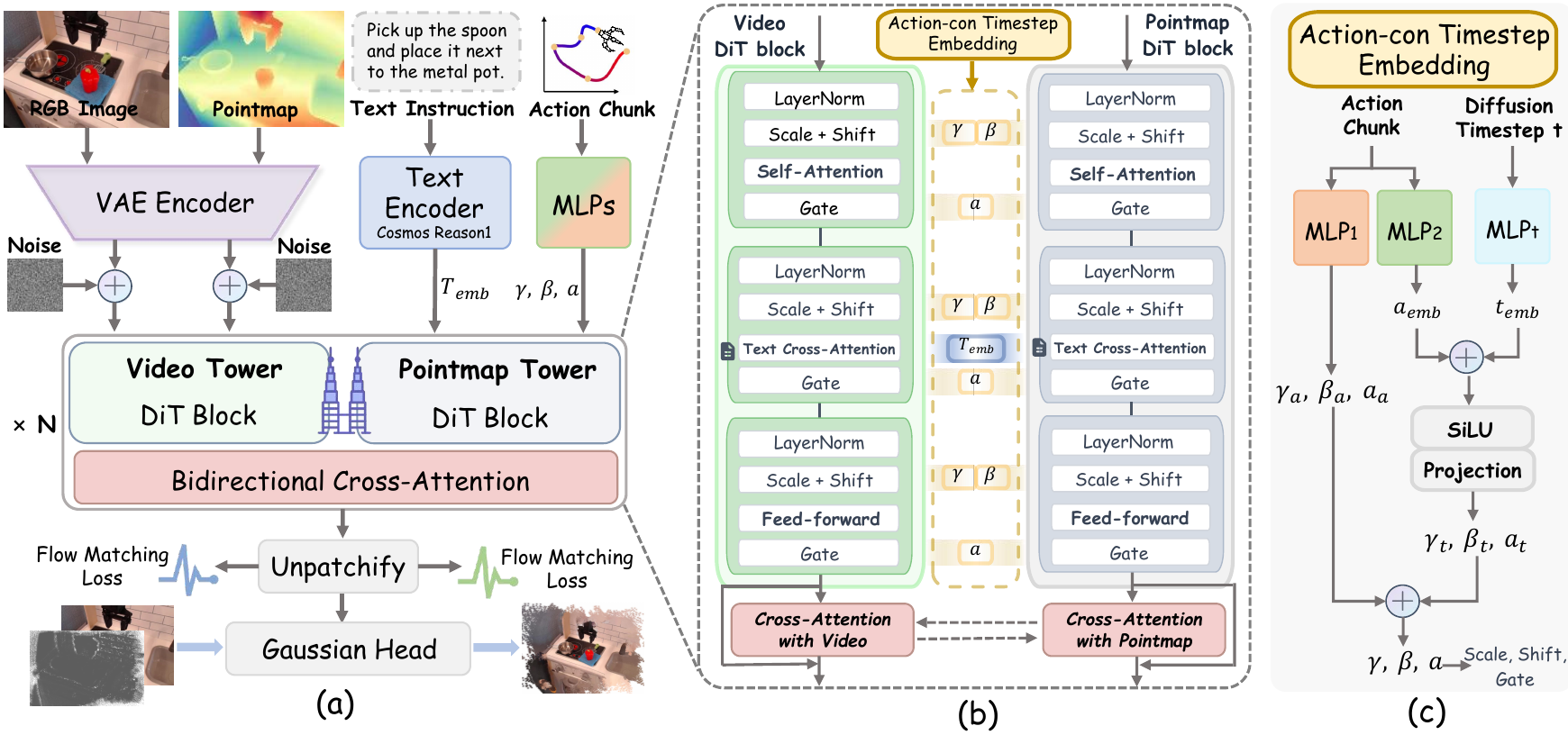}

  \caption{
RoboStereo Architecture.
Symmetric dual DiT towers (a) process RGB and XYZ pointmaps via bidirectional cross-attention for visual-geometric fusion (b) and a Gaussian head for flexible-viewpoint rendering. 
Dual-path action-conditioned timestep embedding mechanism (c) ensures precise frame-level trajectory control.
} 
\label{fig:arch}
   \vspace{-1.5em}
\end{figure*}

\vspace{-1.5em}
\section{Related Work}
\vspace{-0.5em}
\subsection{Embodied World Models and Geometric Consistency}
\vspace{-0.25em}
The development of high-fidelity visual world models has enabled embodied AI policies to learn from realistic future predictions. Early foundation models, such as Cosmos~2.5~\cite{Cosmos2025}, established a baseline for Sim2Real applications by synthesizing diverse video sequences from text and image prompts. However, standard 2D video diffusion models often rely on statistical patterns and lack explicit physical grounding~\cite{team2025inferix,zo3t,jin2024alignment,kong2024hunyuanvideo}. This frequently leads to geometric hallucinations, including object teleportation or structural morphing~\cite{11220316}, which impede long-horizon manipulation planning. To address these issues, recent research focuses on enforcing 3D and 4D structural constraints. FantasyWorld~\cite{FantasyWorld2025} incorporates a trainable geometric branch with dual-pathway processing to maintain multi-view coherence. Similarly, TesserAct~\cite{TesserAct2025} and WristWorld~\cite{WristWorld2025} utilize 4D representations, such as RGB-Depth-Normal streams or spatial projections, to improve consistency in dynamic sequences. The Gaussian World Model (GWM)~\cite{GWM2025} further integrates 3D Gaussian Splatting with latent diffusion for contact-rich predictions. RoboStereo builds on these principles through a Symmetric Dual-Tower DiT architecture. Unlike asymmetric methods, our framework reinforces RGB semantics with continuous 3D pointmap constraints via bidirectional cross-attention, maintaining spatiotemporal coherence and pixel-level 3D grounding.


\vspace{-0.5em}
\subsection{Policy Optimization and Test-Time Scaling for VLAs}
\vspace{-0.25em}
Vision-Language-Action (VLA) models often lack the causal understanding necessary for self-correction when trained solely via supervised imitation learning. To overcome this limitation, world models serve as scalable environments and synthetic data engines for Reinforcement Learning~\cite{dpo,marl-snake,samr1,grpo} (RL). For instance, GigaWorld-0~\cite{gigaworld} demonstrates the potential of world models as a scalable data engine to empower embodied agents. Building on this, the WMPO framework~\cite{WMPO} conducts on-policy Group Relative Policy Optimization (GRPO)~\cite{grpo} within a pixel-space world model, allowing policies to explore and recover from failures. To address sparse reward signals, TGRPO~\cite{TGRPO2025} utilizes LLM-guided reward functions and trajectory-level advantage estimation. RoboStereo extends world-model-based optimization through a framework spanning training to inference. During training, we introduce Imitative-Evolutionary Policy Learning (IEPL) and Open-Exploration Policy Learning (OEPL). Instead of binary rewards, IEPL uses the 4D rendering capabilities of RoboStereo to compute dense, frame-by-frame perceptual distances (LPIPS) from multiple viewpoints, combining the sample efficiency of imitation with the exploratory power of GRPO. Additionally, we introduce Test-Time Policy Augmentation (TTPA) to improve robotic precision. By using RoboStereo as a predictive safeguard, TTPA filters hazardous actions via imagined verification before physical execution.

\vspace{-0.5em}
\section{Method}
\vspace{-0.25em}
Featuring dual video and pointmap towers, RoboStereo processes an initial RGB image and XYZ pointmap alongside a robot action sequence to generate high-fidelity 4D (3D+time) future trajectory chunks, thereby serving as a powerful simulator for VLA policy optimization. We detail the RoboStereo architecture in Sec.~\ref{sec:architecture}, its training strategy in Sec.~\ref{sec:train}, and the unified world model-based policy optimization framework built upon it in Sec.~\ref{sec:policy-opt}.

\vspace{-0.5em}
\subsection{Architecture Design}
\label{sec:architecture}
\vspace{-0.25em}
\noindent\textbf{Symmetric Dual-Tower DiT Framework.} RoboStereo employs a symmetric dual-tower Diffusion Transformer (DiT)~\cite{DiT} that processes RGB video and 3D pointmaps (3-channel XYZ coordinate videos) as joint modalities (Fig.~\ref{fig:arch}(a)). Both towers utilize the Cosmos 2.5 backbone~\cite{cosmos}, sharing identical latent dimensions to enable homogeneous Video Autoencoder (VAE) encoding without additional alignment layers. Each DiT features paired cross-attention layers that facilitate bidirectional information flow (Fig.~\ref{fig:arch}(b)): the video branch queries geometric constraints from the pointmap tower, while the pointmap branch retrieves semantic context from the video tower. This design preserves pretrained feature spaces established during single-modality pretraining while propagating synchronization cues throughout the network. Finally, a 4D Gaussian head~\cite{DA3} projects the generated RGB-XYZ pairs into dynamic Gaussian Splats, enabling flexible novel-view synthesis.

\noindent\textbf{Frame-Level Action Control.} 
To achieve precise action control, we encode actions as continuous vectors and inject them into the DiT backbone via a dual-path action-conditioned timestep embedding mechanism (Fig.~\ref{fig:arch}(c)). An action vector $\mathbf{a}$ is processed by two independent MLPs: one implicitly fuses with the diffusion timestep embedding $e_t$ to generate a base conditioning latent $\mathbf{c} = \text{SiLU}(\mathbf{t}_{\text{emb}} + \text{MLP}_1(\mathbf{a}))$, while the other explicitly predicts modulation offsets $(\gamma_a, \beta_a, \alpha_a) = \text{MLP}_2(\mathbf{a})$. These offsets are additively combined with timestep-derived parameters to form final Adaptive Layer Normalization (AdaLN)~\cite{adaln} coefficients: $\gamma = \gamma_t(\mathbf{c}) + \gamma_a$, $\beta = \beta_t(\mathbf{c}) + \beta_a$, $\alpha = \alpha_t(\mathbf{c}) + \alpha_a$. This explicit modulation enforces frame-level structural guidance at each denoising step, ensuring generated trajectories faithfully follow action sequences. The feature update within a Transformer block is:
\begin{small}
\begin{equation}
\mathbf{x} \leftarrow \mathbf{x} + \alpha \cdot \text{Block}\left(\text{Norm}(\mathbf{x}) \cdot (1 + \gamma) + \beta\right).
\end{equation}
\end{small}

\vspace{-0.5em}
\subsection{Training Strategy for World Model}
\label{sec:train}
\vspace{-0.25em}
Our training follows a two-stage paradigm: single-modality pretraining followed by joint fine-tuning. Both branches are initialized from Cosmos 2.5 pretrained weights, first independently fine-tuned on modality-specific data, then jointly optimized for cross-modal collaboration.

\noindent\textbf{Data Construction.} 
We construct a large-scale 4D robotic manipulation dataset based on Bridge V2~\cite{bridge}, comprising approximately 20,000 third-person videos capturing robotic arms performing diverse kitchen tasks. Videos are standardized to $320 \times 256$ resolution at 5 FPS. Each frame is paired with a 7D action vector $\mathbf{a} = (\Delta x, \Delta y, \Delta z, \Delta\theta_r, \Delta\theta_p, \Delta\theta_y, w_{\text{gripper}})$ specifying gripper displacement, rotation, and width. We employ DepthAnything V3~\cite{DA3} to predict depth maps and construct XYZ pointmaps via inverse perspective projection, forming 3-channel images that maintain pixel-level correspondence with RGB frames. This unified representation enables paired training of both video and geometric streams. More details are provided in the \textbf{Supplementary Material}.

\vspace{0.5em}
\noindent\textbf{{Independent Training.}}
In the first stage, both towers are initialized with Cosmos 2.5 weights and independently fine-tuned to adapt to robotic domains. RGB videos and pointmaps are encoded by VAEs into clean latent representations $\mathbf{z}_v, \mathbf{z}_p$. We sample a timestep $t \sim \mathcal{U}[0,1]$ and independent Gaussian noise $\boldsymbol{\epsilon}_v, \boldsymbol{\epsilon}_p \sim \mathcal{N}(\mathbf{0}, \mathbf{I})$ to construct noisy latents via Rectified Flow~\cite{flow}:
\begin{small}
\begin{equation}
\mathbf{z}^t_m = t\mathbf{z}_m + (1-t)\boldsymbol{\epsilon}_m, \quad m \in \{v, p\}. \label{eq:flow_independent}
\end{equation}
\end{small}
The noisy latents $\mathbf{z}^t_m$, conditioned on the initial frame latents $\mathbf{z}^{(0)}_v$ and $\mathbf{z}^{(0)}_p$, text instruction $\mathbf{c}_{\text{text}}$, and action sequence $\mathbf{c}_{\text{action}}$, are fed into their respective DiT towers to predict the velocity fields:
\begin{small}
\begin{equation}
\mathbf{v}_m^t = \text{DiT}_m(\mathbf{z}_m^t, t, \mathbf{z}^{(0)}_m, \mathbf{c}_{\text{text}}, \mathbf{c}_{\text{action}}),\quad m \in \{v, p\}. 
\label{eq:dit_independent}
\end{equation}
\end{small}
Each model is trained to match the target velocity $\mathbf{z}_m - \boldsymbol{\epsilon}_m$ via MSE loss:
\begin{small}
\begin{equation}
\mathcal{L}_m = \mathbb{E}_{t,\boldsymbol{\epsilon}} \left[\|\mathbf{v}^t_m - (\mathbf{z}_m - \boldsymbol{\epsilon}_m)\|^2_2\right],\quad m \in \{v, p\}. \label{eq:loss_independent}
\end{equation}
\end{small}
This stage enables each branch to strictly capture temporal coherence in video generation and geometric accuracy in pointmap prediction.

\vspace{0.5em}
\noindent\textbf{{Joint Training.}}
In the second stage, we integrate the pretrained towers. To mitigate memory overhead, we freeze all feed-forward network (FFN) layers, fine-tuning only the self-attention and cross-attention modules to align modalities while preserving pretrained feature spaces~\cite{ovi}. We employ paired data latents $(\mathbf{z}_v, \mathbf{z}_p)$ but inject independent Gaussian noise $(\boldsymbol{\epsilon}_v, \boldsymbol{\epsilon}_p)$ with a shared timestep $t \sim \mathcal{U}[0,1]$:
\begin{small}
\begin{equation}
\mathbf{z}^t_m = t\mathbf{z}_m + (1-t)\boldsymbol{\epsilon}_m, \quad m \in \{v, p\}.
\end{equation}
\end{small}
Each backbone predicts velocity fields conditioned on the current noisy latent, timestep, task conditions, and crucially, the intermediate latent from the other modality via bidirectional cross-attention:
\begin{small}
\begin{equation}
\mathbf{v}^t_v = \text{DiT}_v(\mathbf{z}^t_v, t, \mathbf{z}^{(0)}v, \mathbf{c}_{\text{text}}, \mathbf{c}_{\text{action}}, \mathbf{z}^t_p), \ \ \
\mathbf{v}^t_p = \text{DiT}_p(\mathbf{z}^t_p, t, \mathbf{z}^{(0)}p, \mathbf{c}_{\text{text}},\mathbf{c}_{\text{action}}, \mathbf{z}^t_v).
\end{equation}
\end{small}
The objective is the weighted sum of the two modalities’ flow matching losses:
\begin{small}
\begin{equation}
\mathcal{L}_{\text{total}} = \sum_{m \in \{v,p\}} \lambda_m \cdot \mathbb{E}_{t,\boldsymbol{\epsilon}_m,\mathbf{z}_m}\left[\|\mathbf{v}^t_m - (\mathbf{z}_m - \boldsymbol{\epsilon}_m)\|^2_2\right],
\end{equation}
\end{small}
where $\lambda_v = 0.85$ and $\lambda_p = 0.15$. The use of paired data ensures scene consistency, while the shared timestep $t$ forces both branches to synchronize their denoising progress. This strategy facilitates implicit correspondence learning between visual semantics and geometric structures without explicit alignment layers.


\vspace{-0.5em}
\subsection{World Model-Based Policy Optimization}
\label{sec:policy-opt}
\vspace{-0.25em}
\noindent\textbf{{Task Formulation.}}
We formulate robotic manipulation as a sequential decision-making Markov chain~\cite{MDP} where a policy $\pi_\theta$ interacts with an environment to execute language-specified tasks $\mathbf{g}$. At each timestep $t$, the policy observes visual state $\mathbf{s}_t$ (e.g., RGB image) and outputs action chunk $\mathbf{a}_t \in \mathcal{A}$ conditioned on the goal: $\mathbf{a}_t \sim \pi_\theta(\mathbf{a}_t | \mathbf{s}_t, \mathbf{g})$. The environment transitions to the next state $\mathbf{s}_{t+1} \sim p(\mathbf{s}_{t+1} | \mathbf{s}_t, \mathbf{a}_t)$ according to unknown dynamics $p(\mathbf{s}_{t+1} | \mathbf{s}_t, \mathbf{a}_t)$. A complete trajectory is denoted as $\boldsymbol{\tau} = \{(\mathbf{s}_0, \mathbf{a}_0), \ldots, (\mathbf{s}_T, \mathbf{a}_T)\}$.

RoboStereo functions as a learned world model $p_\phi$ that predicts future visual observations conditioned on the current state and action chunk. Formally, given an initial state $\mathbf{s}_0$ and an action sequence $\{\mathbf{a}_0, \ldots, \mathbf{a}_{T-1}\}$, the world model synthesizes an imagined visual trajectory:
\begin{equation}
\hat{\boldsymbol{\tau}} = \{\hat{\mathbf{s}}_1, \ldots, \hat{\mathbf{s}}_T\}, \quad \text{where } \hat{\mathbf{s}}_{t+1} \sim p_\phi(\mathbf{s}_{t+1} | \hat{\mathbf{s}}_t, \mathbf{a}_t). \label{eq:world_model}
\end{equation}
For long-horizon tasks, we employ an autoregressive generation strategy where the last frame of the generated video chunk serves as the initial observation for both the VLA policy and RoboStereo in the next step. This mechanism enables policy evaluation and optimization entirely within the imagined latent space without any real-world interaction. Building upon this formulation, we propose the first unified framework for world-model-based policy optimization, comprising one test-time augmentation strategy and two policy training paradigms.


\begin{figure*}[t]
  \centering
  \setlength{\abovecaptionskip}{-0.02em}   
   \includegraphics[width=1.0\linewidth]{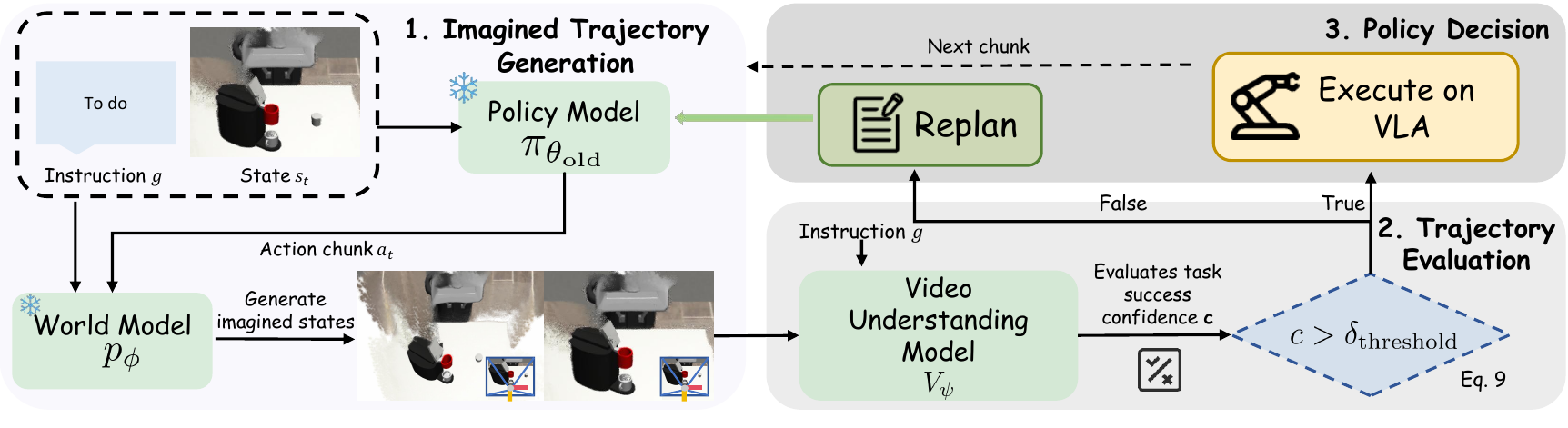}

  \caption{Illustration of Test-Time Policy Augmentation (TTPA).} 
\label{fig:ttpa}
   \vspace{-1.5em}
\end{figure*}

\vspace{0.5em}
\noindent\textbf{{Test-Time Policy Augmentation (TTPA).}}
Direct deployment of VLA in physical environments entails significant safety risks due to potential execution failures. To mitigate this, we introduce TTPA, a high-level planning mechanism that leverages RoboStereo as a mental simulator to verify outcomes before acting.

The workflow proceeds through three integrated stages: (1) {Imagined Trajectory Generation}: Given a candidate action sequence $\mathbf{a}_{t:t+H}$ from the policy, RoboStereo synthesizes a high-fidelity visual forecast $\hat{\boldsymbol{\tau}}$ (Eq.~\ref{eq:world_model}) representing the expected future. (2) {Trajectory Evaluation}: A video understanding model $V_\psi$ assesses the semantic alignment between the imagined trajectory $\hat{\boldsymbol{\tau}}$ and the task goal $\mathbf{g}$, outputting a success confidence score $c = V_\psi(\hat{\boldsymbol{\tau}}, \mathbf{g}) \in [0, 1]$. (3) {Policy Decision}: The system executes the action chunk only if the outcome is deemed safe and successful; otherwise, it triggers a replanning step without physical movement. Formally, the execution logic is defined as:
\begin{small}
\begin{equation}
\text{Decision} = 
\begin{cases}
\text{Execute } \mathbf{a}_{t:t+H}, & \text{if } c > \delta_{\text{safe}}, \\
\text{Replan}, & \text{otherwise},
\end{cases}
\end{equation}
\end{small}
where $\delta_{\text{safe}}$ is a confidence threshold. By simulating actions in RoboStereo’s high-fidelity visual predictions, 
TTPA enables zero-cost failure detection and replanning without physical interaction, reducing real-world trial costs and enhancing 
success rates for complex manipulation tasks.



\vspace{0.5em}
\noindent\textbf{Imitative-Evolutionary Policy Learning (IEPL).}
IEPL exploits the 4D generative capabilities of RoboStereo to establish visual imitation reward signals under flexible camera viewpoints for policy learning. This paradigm is designed to learn from expert demonstrations through trajectory matching via reinforcement learning.

\textbf{\textit{<1> Reward Function Design.}} To guide policy optimization, we formulate a novel 4D visual imitation reward based on the perceptual alignment between expert and policy-induced trajectories within the world model. Given a task $\mathbf{g}$ and expert actions $\mathbf{a}^*_t$, we generate paired 4D imagined rollouts for both the expert actions $\mathbf{a}^*_t$ and the policy actions $\mathbf{a}_t \sim \pi_\theta$, which are rendered into video sequences from identical camera viewpoints:
\begin{small}
\begin{align}
\hat{\boldsymbol{\tau}}_{\text{expert}} &= \{\hat{\mathbf{s}}^*_t\}_{t=1}^T, \quad \text{where } \hat{\mathbf{s}}^*_{t+1} \sim p_\phi(\cdot | \hat{\mathbf{s}}^*_t, \mathbf{a}^*_t), \\
\hat{\boldsymbol{\tau}}_{\text{policy}} &= \{\hat{\mathbf{s}}_t\}_{t=1}^T, \quad \text{where } \hat{\mathbf{s}}_{t+1} \sim p_\phi(\cdot | \hat{\mathbf{s}}_t, \mathbf{a}_t).
\end{align}
\end{small}
The per-step imitation reward is computed as the negative perceptual distance between corresponding imagined frames:
\begin{small}
\begin{equation}
R_{\text{IEPL}}(\boldsymbol{\tau}) = -\sum_{t=1}^{T}\mathcal{D}_{\text{LPIPS}}(\hat{\boldsymbol{\tau}}_{\text{policy}}, \hat{\boldsymbol{\tau}}_{\text{expert}}), \label{eq:iepl_return}
\end{equation}
\end{small}
where $\mathcal{D}_{\text{LPIPS}}$ denotes the well-established perceptual metric LPIPS~\cite{LPIPS}.

\begin{figure*}[t]
  \centering
  \setlength{\abovecaptionskip}{-0.02em}   
   \includegraphics[width=1.0\linewidth]{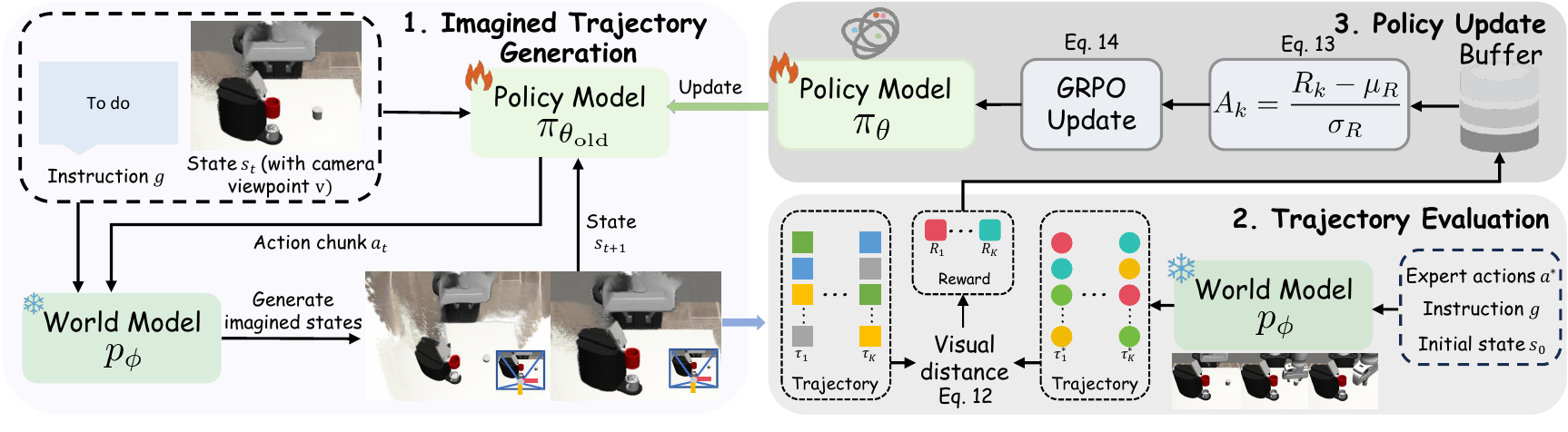}

  \caption{Illustration of Imitative-Evolutionary Policy Learning (IEPL).} 
\label{fig:iepl}
   \vspace{-1.5em}
\end{figure*}

\textbf{\textit{<2> GRPO Optimization.}}
Policy optimization is performed via GRPO~\cite{grpo}. In each iteration, we sample $K$ trajectories $\{\boldsymbol{\tau}^{(k)}\}_{k=1}^K$ from the current policy $\pi_\theta$ and compute their cumulative rewards $\{R_{\text{IEPL}}(\boldsymbol{\tau}^{(k)})\}_{k=1}^K$. The advantages $\hat{A}^{(k)}$ are normalized using the group mean $\mu_R$ and standard deviation $\sigma_R$:
\begin{small}
\begin{equation}
\hat{A}^{(k)} = \frac{R_{\text{\tiny IEPL}}(\boldsymbol{\tau}^{(k)}) - \mu_R}{\sigma_R},  \mu_R = \frac{1}{K}\sum_{k=1}^K R_{\text{\tiny IEPL}}(\boldsymbol{\tau}^{(k)}), \sigma_R = \sqrt{\frac{1}{K}\sum_{k=1}^K \left(R_{\text{\tiny IEPL}}(\boldsymbol{\tau}^{(k)}) - \mu_R\right)^2}.
\label{eq:iepl_adv}
\end{equation}
\end{small}
The policy is then updated by minimizing the clipped surrogate objective:
\begin{small}
\begin{equation}
\mathcal{L}(\theta) = \mathbb{E}_{\boldsymbol{\tau}^{(k)}} \left[ \min\left( \rho_k(\theta) \hat{A}^{(k)}, \; \text{clip}(\rho_k(\theta), 1-\epsilon, 1+\epsilon) \hat{A}^{(k)} \right) - \lambda_{\text{KL}} D_{\text{KL}}(\pi_\theta \| \pi_{\text{ref}}) \right],
\label{eq:iepl_loss}
\end{equation}
\end{small}
where $\rho_k(\theta) = \prod_{t=0}^{T-1} \frac{\pi_\theta(\mathbf{a}_t^{(k)} | \mathbf{s}_t^{(k)}, \mathbf{g})}{\pi_{\theta_{\text{old}}}(\mathbf{a}_t^{(k)} | \mathbf{s}_t^{(k)}, \mathbf{g})}$ denotes the importance sampling ratio. The hyperparameter $\epsilon$ restricts the policy update size, while $\lambda_{\text{KL}}$ weights the KL divergence penalty against a reference policy $\pi_{\text{ref}}$ to prevent mode collapse during open exploration.

\textbf{\textit{<3> Iterative Rollout Process.}} IEPL iterates through three stages: (1) {Imagined Trajectory Generation}: The current VLA policy $\pi_{\theta}$ and expert actions are executed in RoboStereo to generate paired imagined trajectories, $\hat{\boldsymbol{\tau}}_{\text{policy}}$ and $\hat{\boldsymbol{\tau}}_{\text{expert}}$. Every $k$-th frame ($k=8$) from $\hat{\boldsymbol{\tau}}_{\text{policy}}$ is fed back as updated observations, creating closed-loop imitation. (2) {Trajectory Evaluation}: $K$ sampled trajectories under specific camera viewpoints are evaluated based on their perceptual distance to expert rollouts (Eq.~\eqref{eq:iepl_return}), yielding dense imitation rewards. (3) {Policy Update}: The policy is optimized via GRPO (Eq.~\eqref{eq:iepl_loss}), iteratively aligning its behavior with the expert driven by visual reward signals.

IEPL’s innovative visual optimization objective aligns with the large-scale video interaction data used in VLA pre-training, which can maximally activates the pre-training knowledge priors compared to traditional RL algorithms operating in abstract spaces~\cite{WMPO}. The perceptual distance metric provides dense, differentiable rewards that capture nuanced task semantics beyond sparse binary success metrics for effective learning. This approach combines the sample efficiency of imitation learning with the adaptability of reinforcement learning.


\begin{figure*}[t]
  \centering
  \setlength{\abovecaptionskip}{-0.02em}   
   \includegraphics[width=1.0\linewidth]{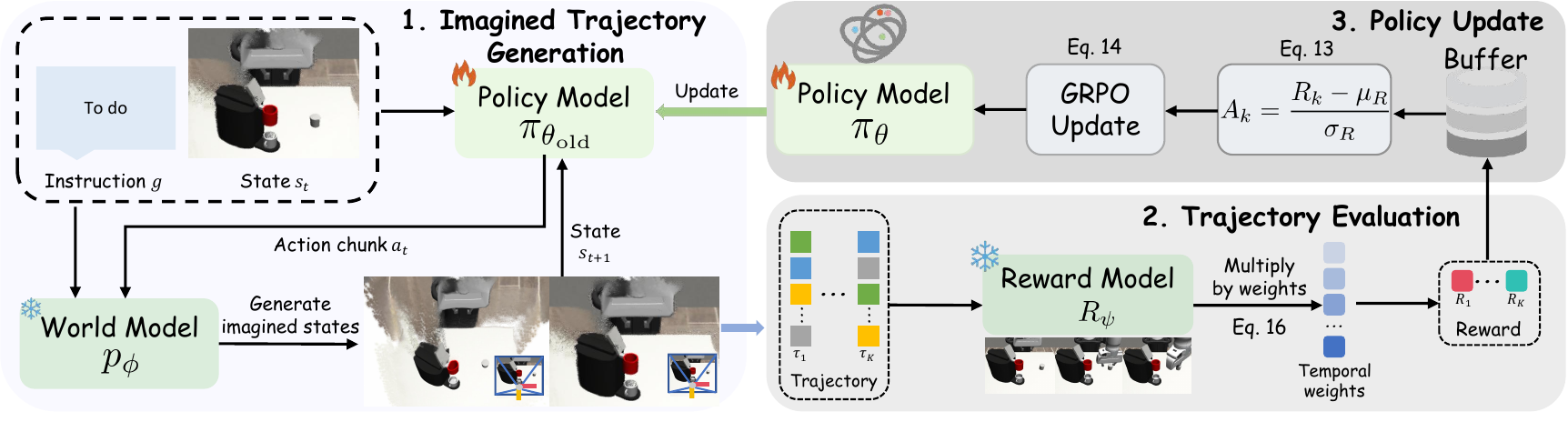}

  \caption{Illustration of Open-Exploration Policy Learning (OEPL).} 
\label{fig:oepl}
   \vspace{-1.5em}
\end{figure*}

\vspace{0.5em}
\noindent\textbf{Open-Exploration Policy Learning (OEPL).}
OEPL leverages RoboStereo for open-ended exploration without expert demonstrations. Unlike IEPL's imitation-focused paradigm, OEPL emphasizes autonomous skill discovery and self-correction through discriminator-based evaluation in open task spaces.

\textbf{\textit{<1> Reward Function Design.}} OEPL employs a video understanding model as reward model $R_\psi$, which is trained to predict task completion likelihood from visual clips. For a sampled trajectory $\boldsymbol{\tau} = \{(\mathbf{s}_t, \mathbf{a}_t)\}_{t=0}^T$, we generate the imagined visual trajectory $\hat{\boldsymbol{\tau}} = \{\hat{\mathbf{s}}_t\}_{t=1}^T$ in RoboStereo via Eq.~\eqref{eq:world_model}. The trajectory is segmented into $N$ overlapping clips $\{\mathbf{c}_i\}_{i=1}^N$, each containing $L$ consecutive frames. The reward model predicts a success confidence for each clip:
\begin{small}
\begin{equation}
r_i = R_\psi(\mathbf{c}_i, \mathbf{g}) \in [0, 1], \quad \mathbf{c}_i = \{\hat{\mathbf{s}}_{t_i}, \ldots, \hat{\mathbf{s}}_{t_i+L-1}\}. \label{eq:oepl_clip_reward}
\end{equation}
\end{small}
The task-level cumulative reward aggregates these clip-level predictions using a temporally increasing weight:
\begin{small}
\begin{equation}
R_{\text{OEPL}}(\boldsymbol{\tau}) = \sum_{i=1}^N w_i \cdot r_i, \quad w_i = \frac{\exp(\beta \cdot i)}{\sum_{j=1}^N \exp(\beta \cdot j)}, \label{eq:oepl_return}
\end{equation}
\end{small}
where $w_i$ assigns higher weights to later clips via temperature $\beta > 0$, encouraging policies to progress toward task completion rather than merely starting correctly.


\textbf{\textit{<2> GRPO Optimization.}} 
Similar to IEPL, we optimize the policy via GRPO. In each iteration, we sample $K$ trajectories $\{\boldsymbol{\tau}^{(k)}\}_{k=1}^K$ from the current policy and compute their cumulative rewards $\{R_{\text{OEPL}}(\boldsymbol{\tau}^{(k)})\}_{k=1}^K$. The advantages $\hat{A}^{(k)}$ are normalized using the group mean $\mu_R$ and standard deviation $\sigma_R$ following Eq.~\ref{eq:iepl_adv}. The policy is then updated by minimizing the optimization objective in Eq.~\ref{eq:iepl_loss}.


\textbf{\textit{<3> Iterative Rollout Process.}} OEPL iterates through three stages: (1) Imagined Trajectory Generation: The VLA policy $\pi_{\theta}$ predicts action sequences from current observations, and RoboStereo generates imagined visual trajectories conditioned on the action chunk. Every $k$-th frame is fed back to the policy as updated observations, creating closed-loop exploration. (2) {Trajectory Evaluation}: $K$ trajectories are sampled and evaluated by the reward model via Eq.~\eqref{eq:oepl_return}, providing reward signals. (3) {Policy Update}: The policy is optimized via Eq.~\eqref{eq:iepl_loss} to maximize rewards through autonomous exploration and self-correction.

OEPL utilizes world models as safe exploration environments, decoupling policy optimization from physical interaction to enable learning from failures. The discriminator-based reward coupled with a temporal weighting mechanism (Eq.~\eqref{eq:oepl_return}) effectively drives success-oriented open exploration. By facilitating autonomous self-correction and novel skill discovery, which are often constrained in pure imitation learning, this paradigm serves as a robust complement to IEPL.

\begin{figure}[t]
\centering
\scalebox{0.73}{ 
    \begin{minipage}[t]{0.62\textwidth} 
        \begin{algorithm}[H]
            \LinesNotNumbered  
            \footnotesize  
            \SetAlgoLined
            \SetAlgoSkip{} 
            \SetInd{0.3em}{0.6em}
            \caption{IEPL Training}
            \label{alg:iepl}
            \KwIn{Policy $\pi_\theta$, World Model $p_\phi$, Expert demos $\{\boldsymbol{\tau}_{\text{expert}}\}$}
            \KwOut{Optimized policy $\pi_{\theta^*}$}
            \For{iteration $i = 1, 2, \ldots$}{
                \tcp{\scriptsize Imagined Trajectory Generation}
                \For{$k = 1$ to $K$}{
                    Sample $\mathbf{s}_0^{(k)}$, $\boldsymbol{\tau}_{\text{expert}}^{(k)}$, camera viewpoint $\mathbf{v}$\;
                    $\hat{\boldsymbol{\tau}}_{\text{expert}}^{(k)} \gets$ Rollout expert in $p_\phi$ (4D)\;
                    $\hat{\boldsymbol{\tau}}_{\text{policy}}^{(k)} \gets$ Rollout $\pi_{\theta_{\text{old}}}$ in $p_\phi$ (4D)\;
                    Render both trajectories to video from viewpoint $\mathbf{v}$\;
                }
                \tcp{\scriptsize Trajectory Evaluation}
                \For{$k = 1$ to $K$}{
                    $r^{(k)} \gets -\frac{1}{T}\sum_t \mathcal{D}_{\text{percept}}(\hat{\mathbf{s}}_t^{(k)}, \hat{\mathbf{s}}_t^{*,(k)})$\;
                    $R_{\text{IEPL}}^{(k)} \gets \sum_{t=0}^{T-1} \gamma^t r^{(k)}$\;
                }
                \tcp{\scriptsize Policy Update via GRPO}
                $\mu_R, \sigma_R \gets$ Compute group statistics\;
                $\hat{A}^{(k)} \gets \frac{R_{\text{IEPL}}^{(k)} - \mu_R}{\sigma_R}$\;
                Update $\theta$ via Eq.~\eqref{eq:iepl_loss}\;
            }
            \Return{$\pi_{\theta^*}$}
        \end{algorithm}
    \end{minipage}
}
\hfill 
\scalebox{0.70}{
    \begin{minipage}[t]{0.65\textwidth}
        \begin{algorithm}[H]
            \LinesNotNumbered  
            \footnotesize
            \SetAlgoLined
            \SetAlgoSkip{}
            \SetInd{0.3em}{0.6em}
            \caption{OEPL Training}
            \label{alg:oepl}
            \KwIn{Policy $\pi_\theta$, World Model $p_\phi$, Reward Model $R_\psi$}
            \KwOut{Optimized policy $\pi_{\theta^*}$}
            \For{iteration $i = 1, 2, \ldots$}{
                \tcp{\scriptsize Imagined Trajectory Generation}
                \For{$k = 1$ to $K$}{
                    Sample $\mathbf{s}_0^{(k)}$\;
                    \For{$t = 0$ to $T-1$}{
                        $\mathbf{a}_t^{(k)} \sim \pi_{\theta_{\text{old}}}(\mathbf{s}_t^{(k)}, \mathbf{g})$\;
                        $\hat{\mathbf{s}}_{t+1}^{(k)} \sim p_\phi(\cdot | \mathbf{s}_t^{(k)}, \mathbf{a}_t^{(k)})$\;
                    }
                }
                \tcp{\scriptsize Trajectory Evaluation}
                \For{$k = 1$ to $K$}{
                    Segment $\hat{\boldsymbol{\tau}}^{(k)}$ into clips $\{\mathbf{c}_i^{(k)}\}$\;
                    $r_i^{(k)} \gets R_\psi(\mathbf{c}_i^{(k)}, \mathbf{g})$ for all $i$\;
                    $R_{\text{OEPL}}^{(k)} \gets \sum_{i=1}^N w_i \cdot r_i^{(k)}$\;
                }
                \tcp{\scriptsize Policy Update via GRPO}
                $\mu_R, \sigma_R \gets$ Compute group statistics\;
                $\hat{A}^{(k)} \gets \frac{R_{\text{OEPL}}^{(k)} - \mu_R}{\sigma_R}$\;
                Update $\theta$ via Eq.~\eqref{eq:iepl_loss}\;
            }
            \Return{$\pi_{\theta^*}$}
        \end{algorithm}
    \end{minipage}
}

\vspace{-1em}
\captionof{figure}{Training procedures for IEPL and OEPL. IEPL (left) learns through visual imitation by matching policy rollouts with expert trajectories. OEPL (right) learns through open exploration guided by a discriminator-based reward model.}
\label{fig:algorithms}
\vspace{-1.5em}
\end{figure}




\vspace{-0.5em}
\section{Experiments}
\vspace{-0.5em}

We systematically evaluate RoboStereo to investigate two core questions: (1) Can RoboStereo achieve highly controllable 4D modeling while preserving spatiotemporal consistency in both visual appearance and geometric structure? (2) Can RoboStereo serve as a high-fidelity embodied simulator to enhance VLA policies within our proposed unified optimization framework, spanning test-time optimization (TTPA) and training-time refinement (IEPL, OEPL)?

\vspace{-0.8em}
\subsection{Experimental Setup}
\vspace{-0.5em}

RoboStereo is initially trained on our 4D embodied manipulation dataset constructed from Bridge V2~\cite{bridge} (Sec.~\ref{sec:train}). Then, we further fine-tune the world model on MimicGen~\cite{MimicGen} simulation data before leveraging it for unified policy optimization. For long-horizon manipulation tasks, RoboStereo sequentially generates 4D scene chunks consisting of 12 frames at a competitive speed of approximately 0.7s per frame. All experiments are performed on a computing cluster equipped with 16 NVIDIA H200 GPUs.

\begin{table}[t]
\centering
\caption{Video evaluation results across visual quality, motion quality and content consistency dimensions. Best results are \textbf{bolded}, and second-best results are \underline{underlined}.}
\vspace{-1em}
\label{tab:video_quality}
\small
\setlength{\tabcolsep}{4pt}
\resizebox{\linewidth}{!}{%
\begin{tabular}{l|ccc|ccc|ccc}
\toprule
\multirow{2}{*}{\textbf{Models}} & \multicolumn{3}{c|}{\textbf{Visual Quality}} & \multicolumn{3}{c|}{\textbf{Motion Quality}} & \multicolumn{3}{c}{\textbf{Content Consistency}} \\
\cmidrule(lr){2-4} \cmidrule(lr){5-7} \cmidrule(lr){8-10}
& Image & Aesthetic & JEPA & Dynamic & Flow & Motion & Subject & Background & Photometric \\
& Quality & Quality & Similarity & Degree & Score & Smoothness & Consistency & Consistency & Consistency \\
\midrule
GigaWorld & \textbf{0.483} & 0.384 & 0.429 & \underline{0.650} & \underline{0.308} & 0.771 & 0.727 & 0.825 & 0.172 \\
Genie & 0.222 & 0.319 & 0.327 & \textbf{0.666} & 0.083 & 0.687 & 0.762 & 0.875 & 0.197 \\
MIND-V & 0.319 & \underline{0.397} & 0.452 & 0.552 & 0.267 & 0.756 & 0.803 & \underline{0.902} & 0.240 \\
RoboMaster & 0.346 & 0.378 & 0.294 & 0.606 & 0.146 & 0.693 & \textbf{0.809} & 0.896 & 0.333 \\
Vidar & 0.408 & \textbf{0.404} & 0.551 & 0.273 & 0.137 & \underline{0.773} & 0.741 & 0.800 & 0.218 \\
Cosmos 2.5 & 0.433 & 0.347 & 0.896 & 0.585 & 0.252 & 0.727 & 0.796 & 0.887 & \underline{0.348} \\
WoW & 0.452 & 0.374 & 0.736 & 0.446 & 0.264 & 0.769 & 0.804 & 0.886 & 0.214 \\
IRASim & 0.339 & 0.357 & \underline{0.909} & 0.412 & 0.204 & 0.685 & 0.806 & 0.891 & 0.342 \\
\midrule
\textbf{Ours} & \underline{0.455} & 0.389 & \textbf{0.912} & 0.610 & \textbf{0.313} & \textbf{0.781} &  \underline{0.807} & \textbf{0.905} & \textbf{0.353} \\
\bottomrule
\end{tabular}
}
\vspace{-1.25em}
\end{table}

\begin{figure*}[t]
  \centering
  \setlength{\abovecaptionskip}{-0.02em}   
   \includegraphics[width=1.0\linewidth]{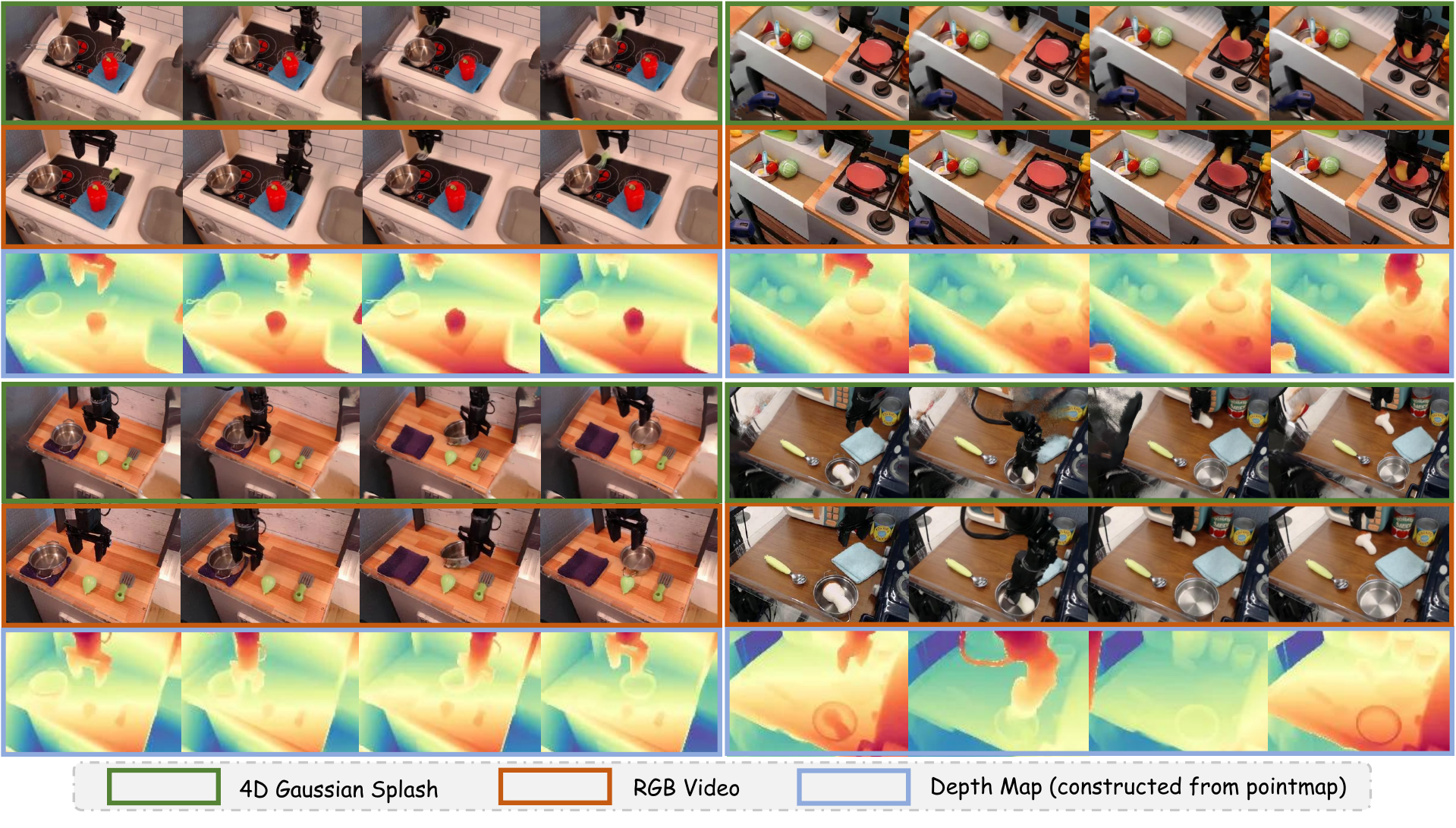}

  \caption{Visualizations of the 4D Gaussian representations, RGB videos, and depth maps produced by RoboStereo. RoboStereo generates precise future trajectories conditioned on action instructions, exhibiting high visual fidelity and geometric consistency.} 
\label{fig:vis}
   \vspace{-1.5em}
\end{figure*}

\vspace{-0.8em}
\subsection{Video Quality Evaluation}
\vspace{-0.25em}

We evaluate RoboStereo's video generation quality against state-of-the-art embodied world models on the WorldArena benchmark~\cite{WorldArena}. The evaluation spans \textbf{16 metrics} across \textbf{six sub-dimensions} (higher is better for all). Detailed metric definitions and implementations appear in the \textbf{supplementary material}.

\vspace{0.5em}
\noindent\textbf{Baselines and Comparative Setup.} 
We compare RoboStereo against eight representative embodied world models: GigaWorld-0~\cite{gigaworld}, Genie Envisioner~\cite{genie}, MIND-V~\cite{mind}, RoboMaster~\cite{robomaster}, Vidar~\cite{vidarc}, Cosmos 2.5~\cite{cosmos}, WoW~\cite{wow}, and IRA-Sim~\cite{IRASim}. For a fair comparison, all baselines with publicly available code are fine-tuned on our Bridge V2-derived dataset following their official configurations.

\vspace{0.5em}
\noindent\textbf{Qualitative and Quantitative Comparison.} 
Tables~\ref{tab:video_quality} and~\ref{tab:video_quality_extended_compact} summarize the video quality quantitative results across six dimensions. RoboStereo ranks first or second on the majority of the 16 metrics, with particularly strong advances in structure- and interaction-centric evaluations. In contrast, although several baselines deliver competitive visual fidelity, they struggle with semantic drift, generating dynamics that diverge from prescribed actions.

Notably, RoboStereo attains top performance across all metrics within the \textbf{\texttt{Physics Adherence}} and  \textbf{\texttt{3D Accuracy}} dimensions, underscoring its exceptional ability to preserve physical plausibility and geometric consistency during intricate interactions. The consistent superiority stems from the interplay of two core designs. (1) The symmetric dual-tower DiT with bidirectional cross-attention enables mutually beneficial feature integration, where geometric constraints from the pointmap tower refine RGB semantics, and visual and motion cues from the RGB tower enhance structural reasoning. (2) Frame-level action conditioning provides fine-grained temporal control, ensuring the generated dynamics strictly adhere to the input actions without temporal drift.

Figure~\ref{fig:vis} further qualitatively demonstrates RoboStereo's capacity to model fine-grained robotic interactions, generating high-fidelity 4D scene evolutions that align closely with ground-truth dynamics. These strengths generalize consistently across both real-world and simulated scenarios, highlighting the model's robust ability to deliver physically realistic and action-accurate 4D generation.

\begin{table}[t]
\centering
\caption{Video evaluation results across physics adherence, 3D accuracy and controllability dimensions. Best results are \textbf{bolded}, and second-best results are \underline{underlined}.}
\vspace{-1em}
\label{tab:video_quality_extended_compact}
\footnotesize
\setlength{\tabcolsep}{3.5pt}
\resizebox{\linewidth}{!}{%
\begin{tabular}{l|cc|cc|ccc}
\toprule
\textbf{Models} 
& \multicolumn{2}{c|}{\textbf{Physics Adherence}} 
& \multicolumn{2}{c|}{\textbf{3D Accuracy}} 
& \multicolumn{3}{c}{\textbf{Controllability}} \\
\cmidrule(lr){2-3} \cmidrule(lr){4-5} \cmidrule(lr){6-8}
& Interaction Quality & Trajectory Acc. 
& Depth Acc. & Perspectivity 
& Instruction Follow. & Semantic Align. & Action Follow. \\
\midrule
GigaWorld 
& 0.523 & 0.152 
& 0.611 & 0.753 
& 0.599 & 0.844 & \textbf{0.109} \\
Genie 
& 0.198 & 0.067 
& 0.844 & 0.513 
& 0.200 & 0.848 & 0.011 \\
MIND-V 
& \textbf{0.570} & 0.137 
& 0.705 & \underline{0.788} 
& 0.593 & 0.865 & 0.079 \\
RoboMaster 
& 0.526 & 0.115 
& 0.801 & 0.731 
& 0.561 & 0.839 & 0.075 \\
Vidar 
& 0.520 & 0.185 
& 0.775 & 0.748 
& 0.573 & 0.856 & 0.039 \\
Cosmos 2.5 
& 0.534 & \underline{0.286} 
& 0.854 & 0.756 
& 0.604 & 0.850 & 0.063 \\
WoW 
& 0.531 & 0.204 
& 0.720 & 0.742 
& 0.565 & \underline{0.873} & 0.042 \\
IRASim 
& 0.560 & 0.257 
& \underline{0.862} & 0.774 
& \underline{0.640} & 0.852 & 0.052 \\
\midrule
\textbf{Ours} 
& \textbf{0.570} & \textbf{0.342} 
& \textbf{0.881} & \textbf{0.797} 
& \textbf{0.723} & \textbf{0.884} & \underline{0.088} \\
\bottomrule
\end{tabular}%
}
\vspace{-1.5em}
\end{table}

\vspace{-0.5em}
\subsection{Policy Optimization Evaluation}
\vspace{-0.25em}
We evaluate RoboStereo's effectiveness as a high-fidelity world model for VLA enhancement within our designed unified policy optimization framework, across three proposed paradigms (Sec.~\ref{sec:policy-opt}): Test-Time Policy Augmentation (TTPA), Imitative-Evolutionary Policy Learning (IEPL), and Open-Exploration Policy Learning (OEPL).

\vspace{0.5em}
\noindent\textbf{Simulation Environment, Base Policy, and Optimization Baselines.}  \\
Evaluations are conducted in the MimicGen simulator~\cite{MimicGen} across three challenging fine-grained manipulation tasks: \texttt{Coffee\_D0} (coffee pod preparation and insertion), \texttt{StackThree\_D0} (stacking three blocks sequentially), and \texttt{Square\_D0} (precise square-peg insertion).

We adopt OpenVLA-OFT~\cite{openvla} fine-tuned via imitation learning on 300 expert demonstrations per task as the base policy $\pi_{\text{base}}$. For simplicity, robot wrist-camera inputs are omitted, following the established protocols in~\cite{WMPO,vlarft}. Using $\pi_{\text{base}}$, we collect 128 execution trajectories per task under varied initial states (encompassing both successes and failures) in MimicGen. These trajectories are utilized to fine-tune RoboStereo for domain alignment and to train a video understanding model (VideoMAE V2~\cite{videomaev2}), which serves as the success evaluator $V_\psi$ for TTPA and reward model $R_\psi$ for OEPL.

To demonstrate the efficacy of our imagined rollouts, we benchmark IEPL and OEPL against two representative optimization baselines: (i) online GRPO~\cite{grpo}, which directly optimizes the policy via simulation interactions using a binary success reward signal, and (ii) offline DPO~\cite{dpo}, which constructs static success-failure pairs from collected offline simulation data. In contrast, IEPL and OEPL execute all optimization procedures within the imagined rollouts generated by RoboStereo. For fair comparison, all methods are strictly constrained to an identical interaction budget of $P=128$ trajectories.

\vspace{0.5em}
\noindent\textbf{Quantitative Results.}
Table~\ref{tab:policy_optimization} reports the average success rate (\%) across 128 diverse evaluation initial states per task. 

\begin{wrapfigure}{r}{0.45\textwidth} 
  \centering
  \vspace{-4em}
  \footnotesize
  \renewcommand{\arraystretch}{1.1}
  \setlength{\tabcolsep}{2.5pt} 
  \captionof{table}{Task success rates (\%) under different optimization paradigms in the MimicGen simulation benchmark.}
  \label{tab:policy_optimization}
  \resizebox{\linewidth}{!}{%
  \begin{tabular}{clcccc} 
  \toprule
  \textbf{ID} & \textbf{Methods} & \textbf{Coffee} & \textbf{StackThree} & \textbf{Square} & \textbf{Mean} \\
  \midrule
  0 & Base policy & 32.6 & 30.2 & 20.2 & 27.7 \\
  \midrule 
  1 & \textbf{TTPA} & 46.7 & 46.9 & 36.2 & 43.3 \\
  \midrule
  2 & DPO & 32.6 & 38.2 & 23.3 & 31.4 \\
  3 & GRPO & 29.8 & 35.1 & 22.7 & 29.2 \\
  4 & \textbf{IEPL} & 47.8 & 44.1 & 37.4 &  43.1 \\
  5 & \textbf{OEPL} & 53.4 & {50.8} & {39.3} & 47.8 \\
  6 & \textbf{Average (1+4+5)} & \textbf{59.8} & \textbf{55.7} & \textbf{46.5} & \textbf{54.0} \\
  \bottomrule
  \end{tabular}%
  }
  \vspace{-2.5em}
\end{wrapfigure}

As reported in Table~\ref{tab:policy_optimization} (ID 1), TTPA delivers a $>56\%$ relative success rate uplift over the base policy. By rejecting low-confidence action sequences through RoboStereo's imagined verification, TTPA substantially enhances deployment reliability without altering the policy weights.

IEPL and OEPL achieve remarkable relative improvements of $>55\%$ and $>72\%$, respectively (Table~\ref{tab:policy_optimization}, IDs 4 and 5), significantly outperforming both online GRPO and offline DPO under the identical rollout budget. These gains are particularly pronounced in contact-rich tasks like \texttt{Square\_D0}, highlighting the innovative designs in training paradigms: (1) IEPL leverages RoboStereo's high-fidelity visual simulation to extract dense visual perceptual rewards for group-relative evolutionary updates. This mechanism maximally activates the VLA's pre-trained visual priors, while semantically rich visual instructions provide fine-grained policy learning guidance. (2) OEPL drives autonomous skill discovery through discriminator-based supervision, fostering robust self-correction capabilities when encountering suboptimal states.

Crucially, integrating all three paradigms (Table~\ref{tab:policy_optimization}, ID 6) yields the most substantial performance gain ($>97\%$ relative uplift), demonstrating profound complementarity among the components. This unified framework represents a holistic approach to policy evolution, encompassing a progression from test-time augmentation to expert imitation to exploratory learning within a single, high-fidelity 4D world model. By shifting the optimization burden from costly, unsafe physical interactions to scalable, imagined rollouts, RoboStereo establishes a highly efficient, closed-loop paradigm that fundamentally unlocks the continuous self-improvement of VLA models.

\vspace{-0.5em}
\subsection{Ablation Study}
\label{ablation}
\vspace{-0.25em}
We conduct a targeted ablation study to assess the impact of two key design choices in RoboStereo: (i) the symmetric dual-tower joint training paradigm (focusing on physics adherence and 3D geometric accuracy), and (ii) the frame-level action conditioning strategy (focusing on controllability and motion quality).

\begin{table}[t!]
\centering
\small
\renewcommand{\arraystretch}{0.9}
\setlength{\tabcolsep}{3.5pt}
\caption{Ablation study conducted on the Bridge dataset. Results demonstrate that joint dual-tower training substantially improves 3D consistency in generated videos, while incorporating action conditioning via timestep embedding modulation yields markedly superior action control and motion quality. Best results are \textbf{bolded}.}
\vspace{-1em}
\label{tab:ablation_study}
\resizebox{0.98\textwidth}{!}{%
\begin{tabular}{l|cc|cc}
\toprule
\multirow{2}{*}{\textbf{Model Variant}} & \multicolumn{2}{c|}{\textbf{Physics Adherence}} & \multicolumn{2}{c}{\textbf{3D Accuracy}} \\
\cmidrule(lr){2-3} \cmidrule(lr){4-5}
& Interaction Quality & Trajectory Acc. & Depth Acc. & Perspectivity  \\
\midrule
(a) Video tower only & 0.538 & 0.320 & 0.821 & 0.758 \\
(b) \textbf{Dual-tower joint training}  & \textbf{0.570} & \textbf{0.342} & \textbf{0.881} & \textbf{0.797} \\
\midrule
\multirow{2}{*}{\textbf{Model Variant}} & \multicolumn{2}{c|}{\textbf{Controllability}} & \multicolumn{2}{c}{\textbf{Motion Quality}} \\
\cmidrule(lr){2-3} \cmidrule(lr){4-5}
& Instruction Follow.  & Action Follow. & Flow Score & Motion Smoothness  \\
\midrule
(c) Action-cond via channel concatenation & 0.622 & 0.061 & 0.238 & 0.703 \\
(d) Action-cond via cross-attention & 0.654 & 0.068 & 0.243 & 0.716 \\
(e) \textbf{Action-cond via timestep embedding}  & \textbf{0.723} & \textbf{0.088} & \textbf{0.313} & \textbf{0.781} \\
\bottomrule
\end{tabular}%
}
\vspace{-1.5em}
\end{table}

As shown in Table~\ref{tab:ablation_study}, compared to isolated RGB-tower training (variant (a)), joint dual-tower training (variant (b)) yields substantial improvements in physics- and geometry-critical metrics (e.g., +5.9\% in Interaction Quality, +7.3\% in Depth Accuracy). Bidirectional cross-attention effectively integrates RGB semantic cues with pointmap-derived 3D constraints, markedly enhancing physical plausibility and long-horizon geometric fidelity. The proposed action-conditioned timestep embedding modulation (variant (e)) achieves superior action alignment and motion coherence, outperforming channel concatenation and cross-attention mechanisms. By injecting action-specific modulation coefficients into diffusion timestep embeddings via extended AdaLN and residual scaling, it attains precise per-frame control without disrupting pretrained features, yielding temporally smooth, physically consistent motion.

\vspace{-1em}
\section{Conclusion}
\vspace{-1em}
We present RoboStereo, a symmetric dual-tower 4D world model addressing critical limitations in geometric consistency and policy optimization for robotic manipulation. Through bidirectional cross-modal enhancement between RGB and pointmap towers, RoboStereo effectively preserves spatiotemporal coherence and alleviates physics hallucinations. Beyond architectural innovations, we establish the first unified world model-based policy optimization framework built on RoboStereo, encompassing Test-Time Policy Augmentation, Imitative-Evolutionary Policy Learning, and Open-Exploration Policy Learning. By shifting the optimization burden from costly physical interactions to scalable imagined rollouts, this work demonstrates that high-fidelity 4D world models can serve as scalable and generalizable foundational infrastructure for the safe, efficient, and continuous improvement of VLAs.

%
%
\bibliographystyle{splncs04}
\bibliography{main}

\clearpage
\setcounter{page}{19}
\begin{center}
    \textbf{\LARGE Supplementary Material}
\end{center}

\section{Inference Efficiency Analysis}

\begin{wrapfigure}{r}{0.42\textwidth} 
  \vspace{-2em} 
  \centering
  \caption{Inference speed comparison.}
  \label{tab:efficiency}
  \vspace{-0.5em}
  \scalebox{0.75}{ 
  \begin{tabular}{lc}
    \toprule
    \textbf{Method} & \textbf{Speed (FPS) $\uparrow$} \\
    \midrule
    WoW~\cite{wow}             & 0.05 \\
    MIND-V~\cite{mind}        & 0.38 \\
    IRASim~\cite{IRASim}       & 0.53 \\
    RoboMaster~\cite{robomaster} & 0.63 \\
    \midrule
    \textbf{Ours (Video Tower)} & \textbf{1.50} \\
    \bottomrule
  \end{tabular}
  } 
  \vspace{-1.5em} 
\end{wrapfigure}
Beyond generation quality, practical embodied simulation requires high inference efficiency to enable large-scale trajectory rollouts. We evaluate the inference speed (measured in frames per second, FPS) of our video generation model against several SOTA baselines. As reported in Table~\ref{tab:efficiency}, RoboStereo achieves an inference speed of 1.50 FPS, maintaining an advantage over baselines. 
This high-efficiency inference is a key enabler for our unified world-model-based policy optimization framework. By facilitating low-latency imagined interactions, RoboStereo enables scalable closed-loop trajectory generation and policy optimization in practice.


\vspace{-0.5em}
\section{Qualitative Analysis of World-Model-Based Policy Optimization Framework}
\vspace{-0.25em}
To further validate the effectiveness of our unified framework, we qualitatively analyze the behavioral patterns of VLA policies optimized via RoboStereo. The proposed optimization paradigms effectively overcome the intrinsic limitations of imitation learning in the base policy, giving rise to emergent recovery behaviors and precise geometric reasoning without requiring real-world interactions.

\vspace{-1.25em}
\begin{figure*}[h]
  \centering
  \setlength{\abovecaptionskip}{-0.02em}   
   \includegraphics[width=1.0\linewidth]{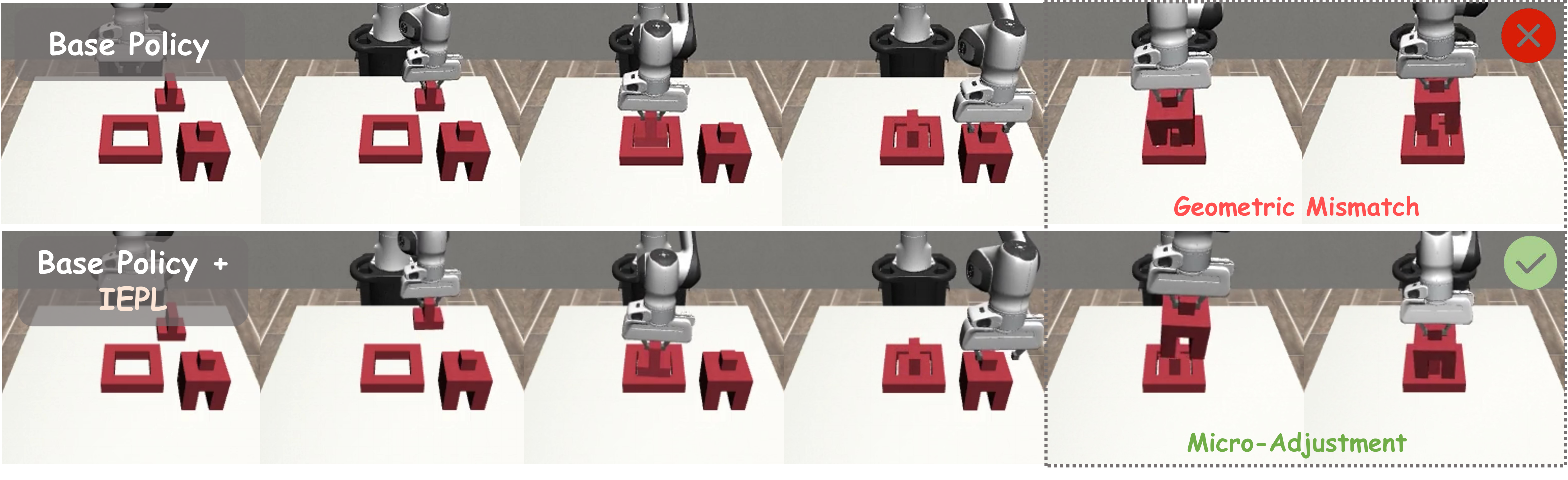}

  \caption{\textbf{Visualization of fine-grained geometric reasoning enabled by IEPL on the \texttt{Three\_Piece\_Assembly\_D0} task.} The base policy (top row) produces coarse actuation and fails to satisfy precise 3D geometric constraints. In contrast, the IEPL-optimized policy (bottom row), guided by dense visual imitation rewards from RoboStereo's imagined trajectories, performs continuous micro-adjustments, achieving accurate part insertion.}

\label{fig:iepl_vis}
   \vspace{-1.5em}
\end{figure*}

\vspace{-0.5em}
\subsection{Emergent Fine-Grained Geometric Reasoning via IEPL}
As illustrated in Fig.~\ref{fig:iepl_vis}, the baseline policy exhibits coarse-grained actuation in interaction-rich insertion tasks. Despite attempting local adjustments, it fails to satisfy the precise 3D geometric constraints between the object and the target receptacle, resulting in persistent misalignment and eventual jamming. This limitation underscores the inadequacy of sparse supervision in conventional imitation learning for acquiring high-precision manipulation skills. By comparison, the policy optimized via IEPL demonstrates markedly superior fine-grained geometric reasoning. Leveraging dense visual imitation rewards synthesized by RoboStereo, the policy acquires the ability to perform smooth, continuous micro-adjustments that precisely align the object's position and orientation. This emergent capability effectively bridges the gap between visual perception and fine-grained motor control, thereby improving success rates on interaction-rich tasks.

\begin{figure*}[t]
  \centering
  \setlength{\abovecaptionskip}{-0.02em}   
   \includegraphics[width=1.0\linewidth]{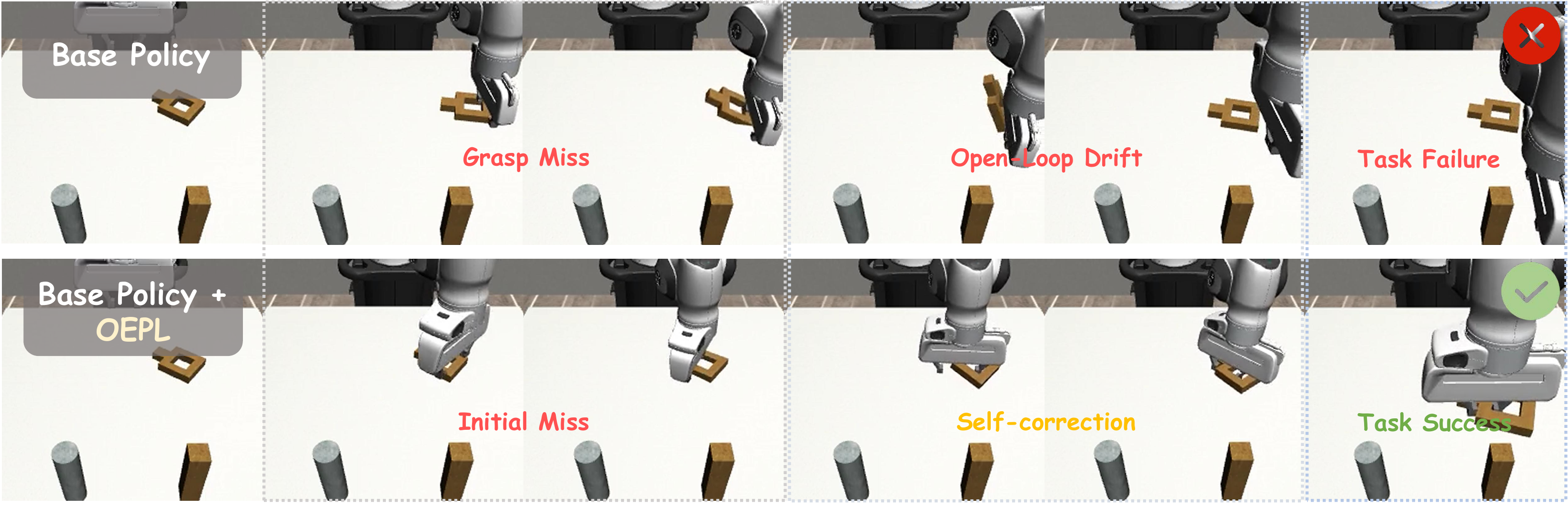}

  \caption{
\textbf{Visualization of emergent self-correction via OEPL on the \texttt{Square\_D0} task.} The base policy (top row) exhibits open-loop failure propagation: after an initial grasp failure, it erroneously treats the failed state as successful and blindly executes the subsequent transport trajectory. In contrast, the OEPL-optimized VLA policy (bottom row) autonomously detects the grasp slip and initiates a corrective re-grasping maneuver, achieving robust task completion.} 
\label{fig:oepl_vis}
   \vspace{-1.5em}
\end{figure*}

\vspace{-0.5em}
\subsection{Emergent Self-Correction via OEPL}
As illustrated in Fig.~\ref{fig:oepl_vis}, the baseline policy exhibits characteristic open-loop failure propagation during grasping tasks. Following an initial grasp failure, it blindly proceeds along the planned transport trajectory without adjustment, proceeds along the planned transport trajectory without adjustment treating an empty gripper as a successful grasp state. This behavior shows the brittleness of standard imitation learning in the presence of execution noise. By comparison, the policy optimized via OEPL exhibits robust autonomous self-correction. Through imagined rollouts augmented by discriminator-based rewards within RoboStereo, the VLA policy acquires the ability to identify suboptimal states and spontaneously triggers a re-grasping maneuver. This emergent self-correction mechanism rectifies execution errors without explicit expert supervision, converting potential task failures into robust task completions.

\begin{figure*}[t]
  \centering
  \setlength{\abovecaptionskip}{2pt}
  \includegraphics[width=1.0\linewidth]{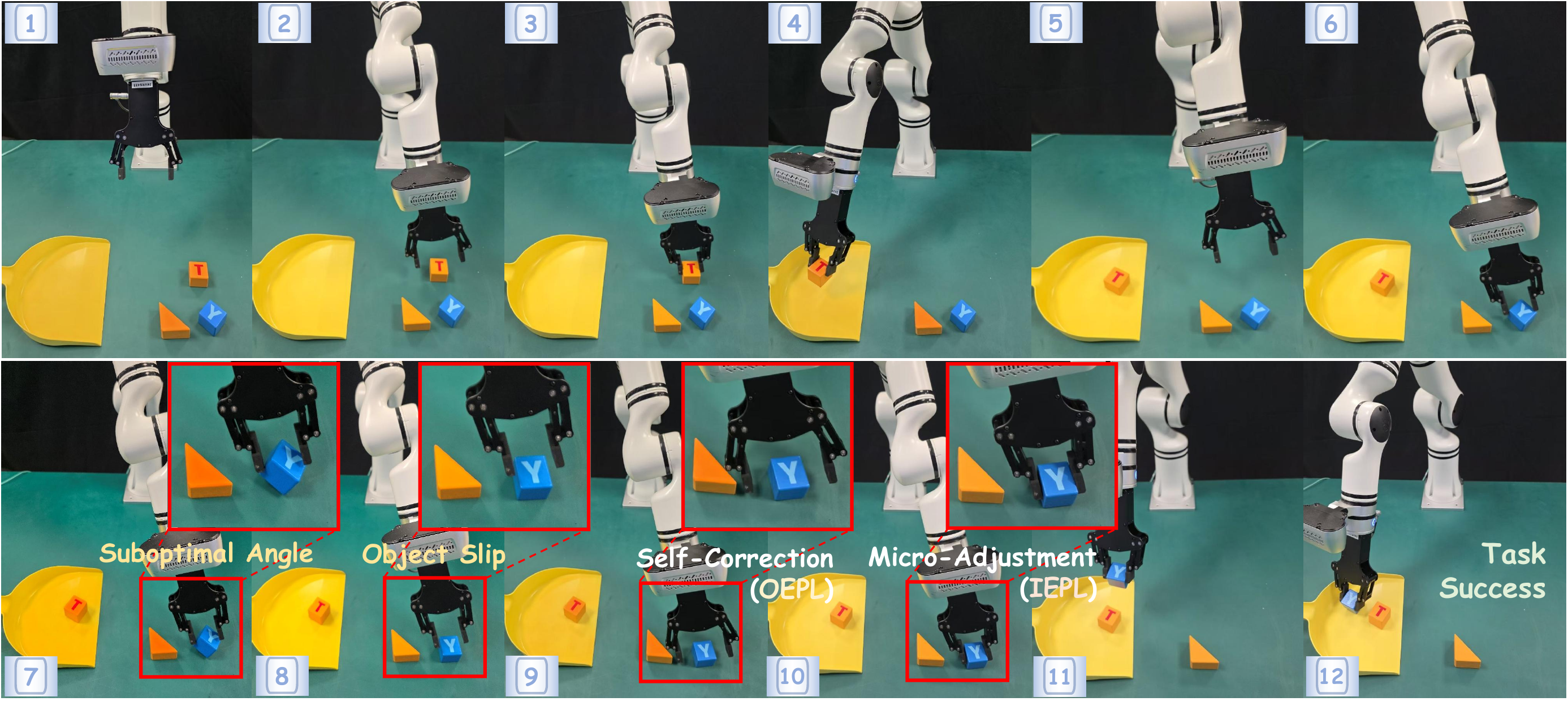}
  \caption{\textbf{Qualitative results of real-world deployment.} The robotic arm is instructed to clear target cubes from a desktop with a prism serving as a distractor. The VLA jointly optimized via IEPL and OEPL exhibits emergent self-correction and fine-grained micro-adjustment capabilities. These critical recovery behaviors absent in the imitation learning baseline.}
  \label{fig:real_world}
  \vspace{-1.5em}
\end{figure*}

\section{Real-World Deployment}
To evaluate the practical applicability of our framework beyond simulated environments, we conduct real-world experiments on a Realman robotic arm using a representative desktop-clearing task. We collect 30 expert demonstration trajectories in the physical setting to fine-tune both the base vision-language-action (VLA) policy (OpenVLA-OFT~\cite{vlarft}) via standard imitation learning (IL) and RoboStereo for domain alignment. The fine-tuned RoboStereo then serves as a high-fidelity imagined rollout environment for IEPL and OEPL optimization under the same hyperparameter configurations as in simulation.

Quantitative evaluation reveals a substantial performance gain: the IL baseline achieves a 30\% success rate (6/20 trials), while joint IEPL and OEPL optimization doubles this to 65\% (13/20 trials). Qualitatively, the IL baseline is highly brittle to execution noise. For example, a suboptimal initial approach angle causes the blue cube to slip, after which the open-loop policy fails to recover and terminates in failure.

In contrast, as shown in Fig.~\ref{fig:real_world}, the jointly optimized VLA policy exhibits two critical emergent capabilities. First, upon detecting grasp slippage, the policy autonomously triggers a replanning and re-grasping sequence, an instance of self-correction enabled by the open-ended state exploration induced by OEPL. Second, during re-grasping, the policy performs precise micro-adjustments to align the end-effector orientation, reflecting the fine-grained geometric reasoning acquired through dense visual imitation rewards provided by IEPL. By tightly integrating imitation-guided refinement with exploratory state-space coverage within a high-fidelity imagined simulator, our approach endows the VLA with robust closed-loop adaptability at a fraction of the cost of extensive real-world trial-and-error.

\section{Data Processing Details}
\label{app:data_processing}

This section details the data preprocessing pipeline for RoboStereo's conditioning inputs. First, we describe the 3D pointmap generation process from RGB videos via monocular depth estimation and inverse perspective projection (Sec.~\ref{subsec:pointmap_gen}). Subsequently, we define the continuous action representation used to condition the world model's dynamics (Sec.~\ref{subsec:action_rep}).

\subsection{3D Pointmap Generation}
\label{subsec:pointmap_gen}
To construct 3D pointmaps, we employ DepthAnything V3~\cite{DA3}, a state-of-the-art video depth estimation model capable of jointly estimating dense depth maps and camera parameters from monocular video sequences. Specifically, for each RGB frame $\mathbf{I} \in \mathbb{R}^{H \times W \times 3}$, DA3 predicts a dense depth map $\mathbf{D} \in \mathbb{R}^{H \times W}$ alongside the camera intrinsic matrix $\mathbf{K} \in \mathbb{R}^{3 \times 3}$ and extrinsic parameters $[\mathbf{R}|\mathbf{t}]$, from which we recover the full 3D geometry. For a given pixel $(u, v)$ with predicted depth $d(u,v)$, its 3D coordinates in the camera coordinate system are computed via inverse perspective projection:
\begin{equation}
\mathbf{p}_{\text{cam}} = d(u,v) \cdot \mathbf{K}^{-1} \begin{bmatrix} u \\ v \\ 1 \end{bmatrix}.
\end{equation}

These camera-space coordinates are subsequently transformed into the world coordinate system. Exploiting the orthogonality of the rotation matrix ($\mathbf{R}^{-1} = \mathbf{R}^\top$), the inverse extrinsic transformation is efficiently computed as:
\begin{equation}
\mathbf{p}_{\text{world}} = \mathbf{R}^\top (\mathbf{p}_{\text{cam}} - \mathbf{t}).
\end{equation}

Finally, the computed 3D coordinates are formatted and stored as a 3-channel image $\mathbf{P} \in \mathbb{R}^{H \times W \times 3}$, where each pixel encodes its corresponding world-space position: $\mathbf{P}(u,v) = \mathbf{p}_{\text{world}}^\top$. This representation guarantees strict pixel-level spatial correspondence with the RGB image $\mathbf{I}(u,v)$, enabling homogeneous processing and aligned latent encoding by the respective Video VAEs.

\subsection{Action Representation}
\label{subsec:action_rep}

To precisely condition the world model, each robot action is formulated as a continuous 7-dimensional vector $\mathbf{a}$ defined in the gripper's local coordinate frame:
\begin{equation}
\mathbf{a} = (\Delta x, \Delta y, \Delta z, \Delta\theta_r, \Delta\theta_p, \Delta\theta_y, w_{\text{gripper}}),
\end{equation}
where:
\begin{itemize}
    \item $(\Delta x, \Delta y, \Delta z)$: Relative translation of the gripper center (in meters).
    \item $(\Delta\theta_r, \Delta\theta_p, \Delta\theta_y)$: Euler angle rotation increments (roll, pitch, yaw, in radians).
    \item $w_{\text{gripper}}$: Gripper width (in meters), indicating the distance between the two fingers.
\end{itemize}
For numerical stability during the diffusion process, all action components are normalized prior to training.

\begin{wraptable}{r}{0.45\textwidth}
  \vspace{-3em}
  \centering
  \caption{Hyperparameters for the GRPO algorithm.}
  \label{tab:grpo_hyperparams}
  \resizebox{0.43\textwidth}{!}{%
    \begin{tabular}{lc}
      \toprule
      \textbf{Hyperparameter} & \textbf{Value} \\
      \midrule
      Optimizer          & AdamW($\beta_1 = 0.9, \beta_2 = 0.999$) \\
      Learning rate      & $5 \times 10^{-6}$ \\
      Training batch size & 64 \\
      Group size $G$     & 8 \\
      Mini-batch size    & 128 \\
      Clip ratio $\epsilon_{low}$  & 0.20 \\
      Clip ratio $\epsilon_{high}$ & 0.28 \\
      Temperature        & 1.6 \\
      \bottomrule
    \end{tabular}%
  }
  \vspace{-2em}
\end{wraptable}

\subsection{Optimization Baseline Details}
We detail the implementation of the optimization baselines, comprising online GRPO and offline DPO. For the GRPO baseline, the key hyperparameters follow the experimental configuration of~\cite{WMPO}, as summarized in Table~\ref{tab:grpo_hyperparams}. The DPO baseline adopts a standard preference-driven offline training paradigm: trajectory data are collected using a supervised fine-tuned OpenVLA-OFT model~\cite{vlarft} to construct a preference dataset, upon which the policy is optimized via the DPO objective. To ensure a fair comparison, the model architecture and optimizer settings are kept identical to those of the GRPO baseline. All training policies are evaluated under the same deployment budget and evaluation protocol.

\section{Architecture Overview}
RoboStereo adopts a symmetric twin backbone design with parallel video and pointmap branches built on an identical diffusion transformer (DiT) architecture. As such, the two backbones share the same number of transformer blocks, heads, head dimensions, and FFNs, enabling symmetry at every layer, as seen in Table~\ref{tab:architecture}

\begin{table}[htbp]
\centering
\caption{Model Architecture Configuration}
\label{tab:architecture}
\resizebox{\textwidth}{!}{%
\footnotesize
\setlength{\tabcolsep}{8pt}
\begin{tabular}{@{} *{7}{>{\centering\arraybackslash}c} @{}}   
\toprule
Model Dim & FFN Dim & Heads & Head Dim & \multicolumn{3}{c}{Blocks} \\
\cmidrule(lr){5-7}
 & & & & Self-Attn & Text Cross-Attn & Tower Cross-Attn \\
\midrule
2048 & 8192 & 16 & 128 & 56 & 56 & 28 \\
\bottomrule
\end{tabular}%
}
\end{table}

\section{Video Quality Evaluation Metrics}
\label{sec:metrics}
To comprehensively assess the perceptual quality of generated videos, we adopt 16 metrics spanning six sub-dimensions from the embodied world model benchmark WorldArena~\cite{WorldArena}. As summarized in Table~\ref{tab:metrics_summary}, these dimensions encompass visual quality, motion quality, content consistency, physics adherence, 3D accuracy, and controllability. To ensure consistent and fair comparison across heterogeneous metrics, all scores are linearly normalized to $[0, 1]$ and direction-aligned such that higher values uniformly denote better performance.

\begin{table}[h]
\centering
\caption{Summary of video quality evaluation metrics in WorldArena~\cite{WorldArena}.}
\vspace{-0.5em}
\label{tab:metrics_summary}
\resizebox{\textwidth}{!}{%
\begin{tabular}{@{}llp{9cm}@{}}
\toprule
\textbf{Dimension} & \textbf{Metric} & \textbf{Computation Method} \\
\midrule
\multirow{8}{*}{\textbf{Visual Quality}}
  & {Image Quality}         & No-reference frame sharpness via MUSIQ~\cite{musiq}; averaged over all frames. \\
  \cmidrule(lr){2-3}
  & Aesthetic Quality     & Per-frame aesthetic score via LAION Aesthetic Predictor~\cite{LAION-AI}; averaged over all frames. \\
  \cmidrule(lr){2-3}
  & \multirow{3}{*}{JEPA Similarity}       & MMD between V-JEPA~\cite{V-jepa} feature distributions of generated and reference videos using a second-order polynomial kernel; exponentiated for normalization. \\
\midrule
\multirow{7}{*}{\textbf{Motion Quality}}
  & Dynamic Degree        & Top-5\% optical flow magnitudes (RAFT~\cite{RAFT}) mapped via sigmoid to quantify salient motion intensity. \\
  \cmidrule(lr){2-3}
  & Flow Score            & Mean optical flow magnitude across all pixels and frames via RAFT; reflects overall motion intensity. \\
  \cmidrule(lr){2-3}
  & \multirow{3}{*}{Motion Smoothness}     & SSIM between VFI-predicted~\cite{vfimamba} and real intermediate frames, weighted by motion magnitude to penalize static sequences. \\
\midrule
\multirow{6}{*}{\textbf{Content Consistency}}
  & \multirow{2}{*}{Subject Consistency}   & Mean cosine similarity of DINO~\cite{dino} frame features relative to first and previous frames; \\
  & & penalized by a dynamic degree to prevent static-video shortcuts. \\
  \cmidrule(lr){2-3}
  & Background Consistency & Mean cosine similarity of CLIP~\cite{clip} frame features relative to first and previous frames; similarly penalized by dynamic degree. \\
  \cmidrule(lr){2-3}
  & \multirow{2}{*}{Photometric Consistency} & Inverse of forward-backward optical flow warp error (AEPE); \\
  & & scaled by dynamic degree to penalize near-static videos. \\
\midrule
\multirow{5}{*}{\textbf{Physics Adherence}}
  & Interaction Quality   & VLM-based (Qwen3-VL~\cite{qwen3}) Likert scoring (1--5) of physical plausibility of robot-object contact, normalized to $[0,1]$. \\
  \cmidrule(lr){2-3}
  & \multirow{3}{*}{Trajectory Accuracy}   & Normalized Dynamic Time Warping (NDTW) between SAM3~\cite{sam3}-extracted arm trajectories and ground-truth; inverted so higher is better. \\
\midrule
\multirow{6}{*}{\textbf{3D Accuracy}}
  & \multirow{3}{*}{Depth Accuracy}        & Absolute relative error between median-scaled monocular depth maps (Depth-Anything~\cite{depth}) of generated and reference videos; inverted for normalization. \\
  \cmidrule(lr){2-3}
  & \multirow{3}{*}{Perspectivity}         & VLM-based (Qwen3-VL) Likert scoring of 3D geometric plausibility, including scale, occlusion, and lighting consistency, normalized to $[0,1]$. \\
\midrule
\multirow{9}{*}{\textbf{Controllability}}
  & \multirow{3}{*}{Instruction Following} & VLM-based (Qwen3-VL) Likert scoring of action-type, target-object, and task-state adherence to instruction, normalized to $[0,1]$. \\
  \cmidrule(lr){2-3}
  & Semantic Alignment    & Cosine similarity between CLIP text encodings of Qwen2.5-VL~\cite{qwen3}-generated descriptions of generated and reference videos. \\
  \cmidrule(lr){2-3}
  & \multirow{3}{*}{Action Following}      & Mean pairwise CLIP feature dissimilarity across videos generated from distinct instructions; measures instruction-driven output diversity. \\
\bottomrule
\end{tabular}%
}
\vspace{-1em}
\end{table}

\end{document}